\documentclass[conference]{IEEEtran}

\usepackage{times}
\usepackage{epsfig}
\usepackage{graphicx}
\usepackage{amsmath}
\usepackage{amssymb}

\usepackage[noend]{algorithmic}
\usepackage{algorithm}
\usepackage{mathtools, nccmath}

\usepackage[pagebackref=false,breaklinks=true,colorlinks,bookmarks=false]{hyperref}

\usepackage{booktabs}

\begin{document}

\title{Two layer Ensemble of Deep Learning Models for Medical Image Segmentation}
\author{\IEEEauthorblockN{Truong Dang, Tien Thanh Nguyen, John McCall, Eyad Elyan, Carlos Francisco Moreno-García}
\IEEEauthorblockA{School of Computing
Robert Gordon University, Aberdeen, UK\\
}}

\maketitle  

\date{September 2020}
\bstctlcite{IEEEexample:BSTcontrol}
\begin{abstract}
In recent years, deep learning has rapidly become a method of choice for the segmentation of  medical images. Deep Neural Network (DNN) architectures such as UNet have achieved state-of-the-art results on many medical datasets. To further improve the performance in the segmentation task, we develop an ensemble system which combines various deep learning architectures. We propose a two-layer ensemble of deep learning models for the segmentation of medical images. The prediction for each training image pixel made by each model in the first layer is used as the augmented data of the training image for the second layer of the ensemble. The prediction of the second layer is then combined by using a weights-based scheme in which each model contributes differently to the combined result. The weights are found by solving linear regression problems. Experiments conducted on two popular medical datasets namely CAMUS and Kvasir-SEG show that the proposed method achieves better results concerning two performance metrics (Dice Coefficient and Hausdorff distance) compared to some well-known benchmark algorithms.

\end{abstract}
\section{Introduction}

Segmentation is the process of partitioning an image into multiple segments to locate objects and boundaries. Before the rise of Deep Neural Networks (DNN), most of the successful segmentation algorithms used hand-crafted features combined with a machine learning classifier such as Random Forest \cite{shotton_2008_random_forest_segmentation} or Support Vector Machine \cite{soatto_2009_svm_segmentation}. Even though subsequent research have achieved noticeable improvements by incorporating richer context information \cite{carreira_semantic_2012} or by applying structured prediction techniques \cite{krahenbuhl_efficient_2011,xuming_2004_multiscale}, the performance of these systems remained limited because the hand-crafted features are not representative enough for real-world usage. With the success of DNNs in image classification in 2012 \cite{krizhevsky_imagenet_2012}, researchers began to apply this new architecture to segmentation. Some notable results in this direction include Fully Connected Networks (FCN) \cite{fcn_paper_2015} and SegNet \cite{segnet_kendall_2017}. 
Applying deep learning techniques to medical imaging has brought many successes, such as the introduction of a novel architecture called Unet and successfully applied it to the segmentation of neuronal structures in electron microscopic stacks \cite{unet_olaf_2015}. This network continues to be widely used for segmentation. Another notable example is in \cite{zhang_deep_2015} which used T1-weighted, T2-weighted, and fractional anisotropy image patches of 13x13 in size as input to a Convolutional Neural Network (CNN) for segmentation of infant brains which are considered to be much more difficult than adult brains. This approach outperforms other commonly used segmentation algorithms when tested on a set of manually segmented isointense stage brain images.  Deep learning methods are highly effective for cases when the dataset is large. For example, the first success in deep learning was a network trained on the ImageNet Large Scale Visual Recognition Challenge (ILSVRC) dataset \cite{krizhevsky_imagenet_2012}, which contained 1 million annotated images. However, medical image datasets are much smaller, usually about 1,000 images\cite{shen_medical_2017}. This creates an important challenge for creating deep medical models which are robust against overfitting. Another problem is that popular optimizers for training deep neural networks such as Stochastic Gradient Descent (SGD) generally require much manual tuning of optimization parameters \cite{quoc_deep_2011}. Despite the fact that there has been some alternative methods which require less parameter tuning, such as Adam  \cite{kingma_adam_2015}, these methods do not generalize as well as SGD \cite{ashia_marginal_2017}. The manual parameter tuning causes a challenge in selecting suitable deep models for a specific problem. Therefore, because medical image analysis requires reliable predictions from automated systems due to its critical nature, it is essential to leverage the strong points of multiple segmentation algorithms to improve on the final results. 

Ensemble learning is a popular technique in which multiple learners are combined to make a collaborated decision. The key challenge is to build an effective ensemble method to combine the results of segmentation algorithms. The paper is organized as follows. In section 2, we briefly review the existing approaches relating to segmentation in medical image analysis and the ensemble learning. In section 3, we propose a novel two-layer ensemble method to combine the results of segmentation algorithms. Because segmentation gives a pixel-level output, the prediction results by the segmentation algorithms are concatenated with the original image as input to segmentation algorithms in the second layer. Dice Coefficient and Hausdorff distance are used as the evaluation metrics. The details of experimental studies on two public datasets are described in section 4. Finally, the conclusion is given in section 5. 

\section{Background and Related Work}

\subsection{Semantic segmentation in medical image analysis}

With the success of \cite{krizhevsky_imagenet_2012} in applying deep Convolutional Neural Network (CNN) to the problem of image classification, deep learning has become the most popular approach in computer vision. Since then, many notable deep architectures have been proposed to solve vision problems. For example, VGG16 \cite{vgg16_paper_2015} was a deep CNN for image classification using a stack of convolution layers with small receptive fields in the first layers instead of few layers with big receptive fields like previous models. This allows the model to have much fewer parameters and more non-linearity, which makes the decision function more discriminative and the model easier to train. VGG16 managed to achieve a top-5 accuracy of 92.7\% on the ImageNet Large Scale Visual Recognition Challenge (ILSVRC)-2013 dataset. Another notable model is ResNet \cite{resnet_paper_2016}, which was motivated by the problem of training a really deep architecture. The network uses shortcut connections in order to perform identity mapping, i.e. instead of learning a function, the layers having shortcut connections learn the residual mapping. This allows Resnet to have a very deep network at 152 layers while achieving 96.4\% accuracy on the ILSVRC-2016 competition. 

Generally, deep image classification models are trained on large datasets, such as ImageNet \cite{imagenet_cvpr_2009} which have around 1 million images. However, in the problem of semantic segmentation, in which a model must predict the class of each pixel in the image, the scale of available datasets is not as large as in image classification \cite{garcia-garcia_survey_2018}. To overcome this limitation, practitioners usually use a pre-trained classification network and finetune it for segmentation. Most deep learning based semantic segmentation architectures are inspired by Fully Convolutional Network (FCN) \cite{fcn_paper_2015}, which creates a segmentation network by using an existing classification network and replace the fully connected layers with convolutional ones to output spatial maps instead of classification scores. Those maps are then upsampled to produce dense pixel-level output. This architecture is considered the cornerstone of deep learning applied to semantic segmentation \cite{garcia-garcia_survey_2018}. Another notable example is DeepLab \cite{deeplab_chen_2018} which makes use of Conditional Random Fields (CRF) \cite{crf_koltun_2013} as a post-processing step for the refinement of the segmentation result. The proposed architecture models each pixel as a node in the random field and employs a fully connected factor graph in which one pairwise term is used for each pixel pair irrespective of their distance. This allows the model to incorporate both short-range and long-range information into account, facilitating the restoration of detailed structures in the segmentation process that was lost due to the spatial invariance of CNN. 

Segmentation is considered one of the most essential medical imaging process as it extracts  the region of interest (ROI) which is then used in clinical applications. Therefore, it has seen the widest variety of proposed methodology, including deep architectures specifically designed to tackle problems in medical image analysis. A notable example is UNet \cite{unet_olaf_2015} which consists of a contracting path and an expanding path designed symmetrically. To help with localization, high resolution features from the contracting path are combined with the upsampled output. An important difference of UNet compared to previous architectures is that the upsampling part also has a large number of feature, channels, which allow the network to propagate context information to higher resolution layers. The network does not have any fully connected layer and therefore can be trained on images of arbitrary size via an overlap-tile strategy. In recent years, Recurrent Neural Networks (RNNs) have also become widely used for medical image segmentation. For example, in \cite{xie_spatial_2016} a spatial clockwork RNN was used to segment perimysium in histopathology images. The authors applied the RNN four times in different orientations in order to incorporate bidirectional information from left/top and right/bottom neighbors. For 3D brain segmentation, \cite{kleesiek_deep_2016} trained a 3D-CNN by using mini-batches of multiple cubes, whose size was larger than the input size. Their proposed model could take an arbitrary-sized 3D patch as input and would output a block of predictions per input, which is similar to FCN. Over four different brain segmentation datasets, their proposed method achieved the highest average specificity measure, with no significant loss in sensitivity. Some researchers have also used graphical models such as Conditional Random Fields as a post-processing step to refine the segmentation results \cite{alansary_fast_2016}. 

\subsection{Ensemble learning}

Ensemble learning is a popular approach in machine learning for combining a collection of classifiers for the collaborative decision. Designing an ensemble system requires two stages, namely ensemble generation and ensemble integration. In the ensemble generation, multiple classifiers are generated by using either a homogeneous strategy (training a learning algorithm on multiple training sets generated from the original training data) \cite{manuel_doweneed_2014, nguyen_projection_2019} or a heterogeneous strategy (training different learning algorithms on the original training data) \cite{nguyen_granular_2018, nguyen_confidence_2020}. A combining method is then used to aggregate the predictions of the constituent classifiers in the ensemble integration stage to obtain the collaborated prediction. Several top-performing methods for classification have been reported including Random Forest \cite{breiman_random_2001}, XgBoost \cite{chen_xgboost_2016}, and Rotation Forest \cite{rodriguez_rotation_2006}.

Recently, there is increasing interest in the ensemble generation inspired by the success of DNNs. Instead of using only one layer like in traditional ensemble models, the ensemble systems were made to train deeply through multiple layers. The first deep ensemble system was proposed by Zhou and Feng \cite{zhou_deep_2017} (called gcForest), containing multiple layers of two Completely-Random Tree Forests and two Random Forests in each layer. Each forest in a layer outputs a class vector, which is then concatenated to the original data as the input data to the next layer. Utkin et al. \cite{utkin_forest_2019} proposed a weighted average approach for gcForest by associating each tree with a weighted vector for its class distribution vector. The optimal weight vectors of each trees in one layer are found by minimizing the distance between the class label vector in a binary encoding scheme and the weighted prediction vector of this forest. The authors proposed to set only a weight vector for each group in order to reduce the computational overhead. Nguyen et al. \cite{thanh_mulilayer_2020} proposed MULES, a deep ensemble system with classifier and feature selection in each layer. The optimal configuration of each layer is found by using a bi-objective optimization problem in which the two objectives to be maximized are classification accuracy and diversity of the ensemble in each layer. Qi et al. \cite{zhiquan_support_2016} introduced a deep ensemble model in which each layer consists of an ensemble of Support Vector Machine (SVM) classifiers \cite{burges_tutorial_1998}. The model parameters, such as the kernel functions of the SVM classifiers, the number of classifiers, and the weights of the features are found by AdaBoost \cite{schapire_adaboost_1997}. 


\section{Proposed Method}

Our proposed method is inspired by multi-layer ensemble learning architectures, in which the segmentation algorithms in one layer train the segmentation models of this layer on the new training data generated by the preceding layer \cite{zhou_deep_2017}. Applied to segmentation of medical images, this facilitates the successive refinement of segmentation results through each layer. It is recognized that the most successful segmentation algorithms in recent years have been based on DNNs \cite{litjens_survey_2017}, and even though deep learning models can be trained in parallel using GPU, a multi-layer ensemble model of deep learning-based segmentation algorithms would require a lot of computational resources. Therefore, an important question arises: How many layers should a deep ensemble model extend? \cite{thanh_mulilayer_2020} showed that on some datasets, the number of layers of multi-layer ensemble obtained was 2 or 3 only. Based on this observation, we introduce a novel two-layer ensemble model for segmentation of medical images. Figure \ref{fig:proposed_method_overview} shows the high-level overview of our proposed method.

\subsection{Two-layer ensemble for segmentation}

Let $\textbf{D}=\{\textbf{I}_n,\textbf{Y}_n\}_{n=1}^{N}$ be the training set where $N$ is the number of images, $\textbf{I}_n$ is an input image of size $(W,H,C)$ with $H$ being the image height, $W$ the image width, and $C$ is the number of channels ($C=1$ for grayscale, $C=3$ for color image). The mask $\textbf{Y}_n$ is also an image of size $(W,H)$, with each entry $\textbf{Y}_n(i,j)(i=1,...,W;j=1,...,H)$ showing which group the pixel $\textbf{I}_n(i,j)$ belongs to, i.e $\textbf{Y}_n(i,j) \in \mathcal{Y}$, where $ \mathcal{Y} = \{y_m\},m=1,...,M$ is the set of all classes and $M$ is the number of classes.

We aim to learn a hypothesis $\textbf{h}: \textbf{I}_n \rightarrow \textbf{Y}_n$ (i.e segmentation model) to approximate the unknown relationship between each image and its corresponding mask, and then use this hypothesis to assign a label for each unsegmented image. We also denote $\{\mathcal{K}_k\}_{k=1}^K$ as the set of $K$ segmentation algorithms. Each segmentation algorithm $\mathcal{K}_k$ learns on $\textbf{D}$ to obtain a trained segmentation model $\textbf{h}_k$. In ensemble learning, we train segmentation algorithms
$\{\mathcal{K}_k\}_{k=1}^{K}$ on $\textbf{D}$ to get $K$ trained segmentation models $\{\textbf{h}_k\}_{k=1}^K$

\begin{figure}
    \begin{center}
        \includegraphics[width=0.4\textwidth ]{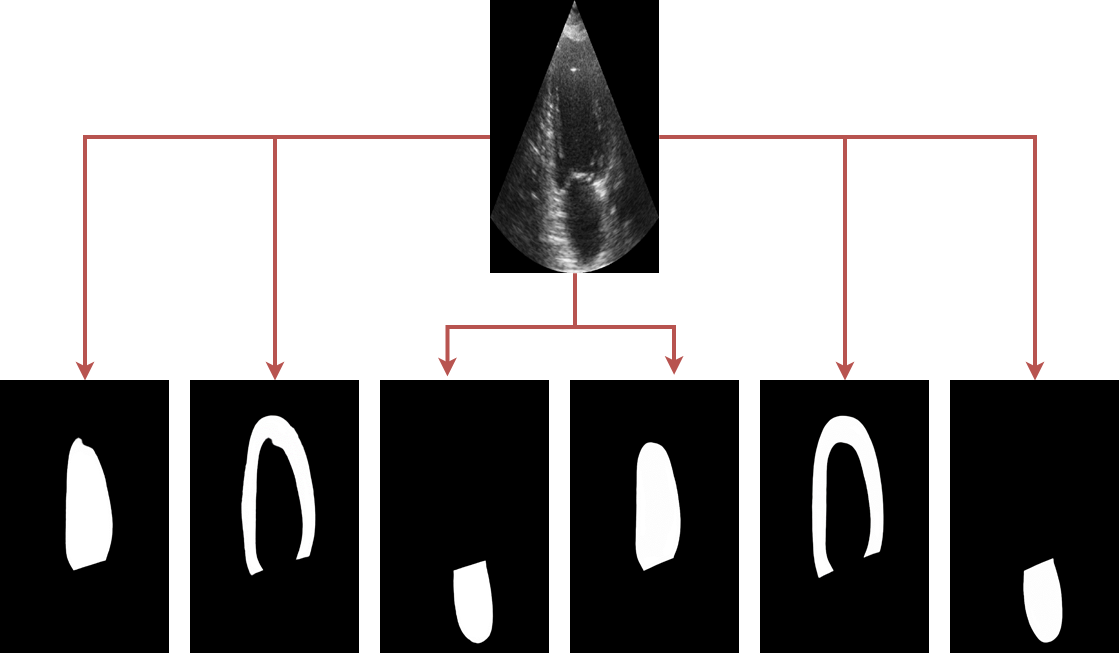}
    \end{center}
    \caption{Example of prediction results on CAMUS dataset. Top: Original image. Bottom is the predictions for Left ventricle, Myocardium and Left atrium classes, made by UNet and LinkNet with backbones ResNet34 and VGG16, respectively. The result has been multiplied by 255 for visualization.}
    \label{fig:metadata_example}
\end{figure}

In the next step, we generate the training data for the second layer of ensemble. 
Based on the results of \cite{thanh_mulilayer_2020} and the stacking generalization model \cite{nguyen_granular_2018}, we propose a two-layer deep ensemble architecture for segmentation in medical image analysis (Figure \ref{fig:proposed_method_overview}). Firstly, the training set $\textbf{D}$ is divided into $T$ disjoint parts $\{\textbf{D}_1, \textbf{D}_2, ..., \textbf{D}_T \}$, where $\textbf{D}=\textbf{D}_1 \cup \textbf{D}_2 \cup ... \cup       \textbf{D}_T,\textbf{D}_{t_1} \cap \textbf{D}_{t_2}= \emptyset (t_1,t_2=1,...,T, t_1 \neq t_2)$. Then for each part $\textbf{D}_t(t=1,...,T)$, the segmentation algorithms $\{\mathcal{K}_k\}_{k=1}^{K}$ will learn on its complementary $\textbf{D} \setminus \textbf{D}_t$ to obtain segmentation models $\textbf{h}_{k,t}$. The images in $\textbf{D}_t$ are then segmented by using these segmentation models.
Let $P_k(y_m|\textbf{I}_n(i,j))$ be probability prediction that $\textbf{h}_{k,t}$ assigns pixel $\textbf{I}_n(i,j)$ to be in class $y_m$. The prediction of $\textbf{h}_{k,t}$ showing the probability all pixels of image $\textbf{I}_n$ belonged to class $y_m$ is given by a matrix:

\begin{figure*}
    \begin{center}
        \includegraphics[width=0.9\textwidth, height=0.3\textwidth]{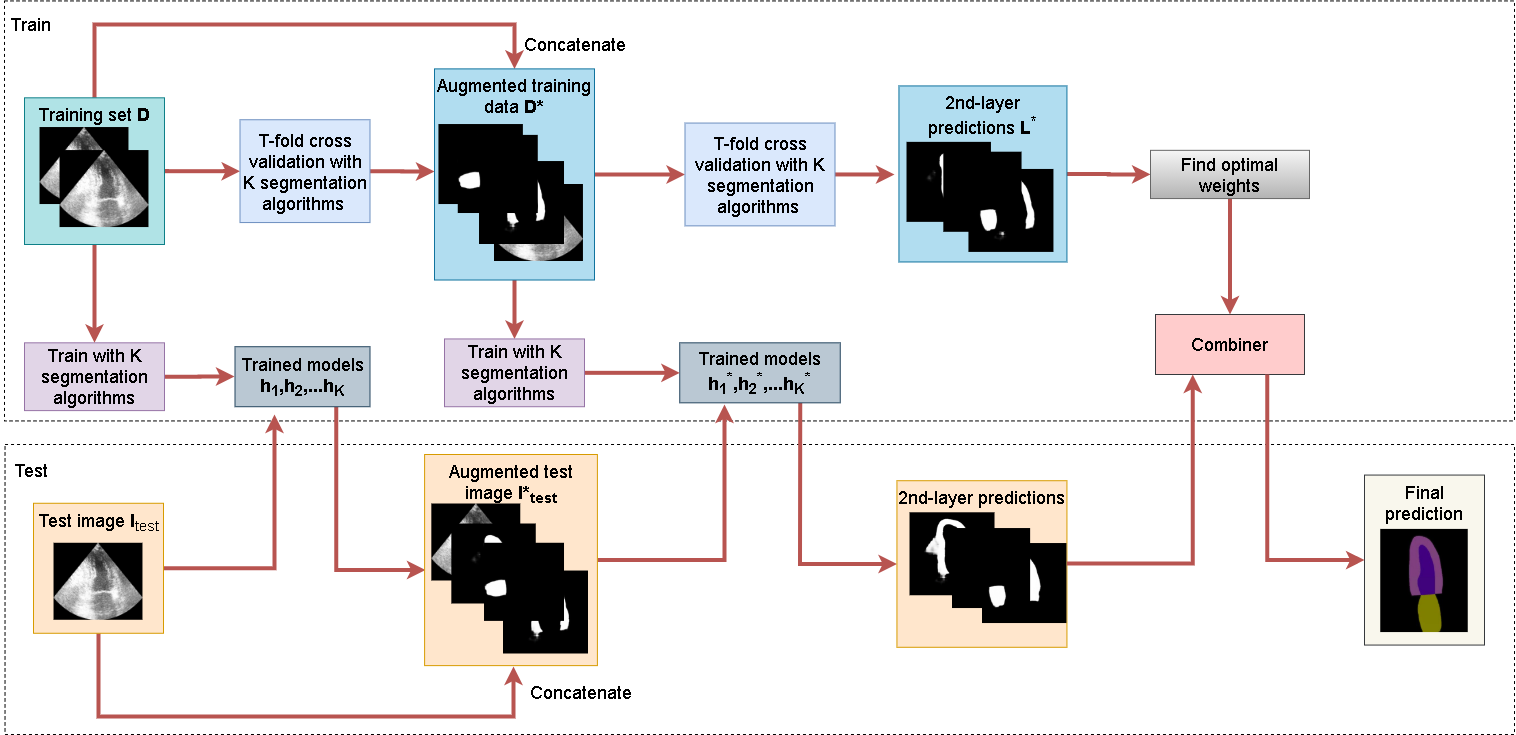}
    \end{center}
    \caption{High-level overview of the proposed method.}
    \label{fig:proposed_method_overview}
\end{figure*}


\begin{equation}
    \resizebox{0.9\columnwidth}{!}{%
    $\mathbf{P_k}(y_m|\textbf{I}_n) = \begin{bmatrix}
        P_k(y_m | \textbf{I}_n(1,1)) & P_k(y_m | \textbf{I}_n(1,2)) & \cdots & P_k(y_m | \textbf{I}_n(1,H)) \\
        \cdots & \cdots & \cdots & \cdots \\
        P_k(y_m | \textbf{I}_n(W,1)) & P_k(y_m | \textbf{I}_n(W,2)) & \cdots & P_k(y_m | \textbf{I}_n(W,H)) \\
    \end{bmatrix}$
    }
\end{equation}
For each image $\textbf{I}_n$, there will be $M \times K$ prediction matrices $\mathbf{P_k}(y_m|\textbf{I}_n)$ illustrated in Figure \ref{fig:metadata_example}. In this study, we propose augment the training data for the second layer of ensemble by concatenating these $M \times K$ prediction matrices to the original training images to create new images $\textbf{I}^{*}_n$. The prediction matrix $\{\mathbf{P_k}(y_m|\textbf{I}_n)\}$ serves as an additional channel of the original image $\textbf{I}_n$. In total, the new images $\textbf{I}^{*}_n$ will have $C+M \times K$ channels:
\begin{equation}
    \label{eq:level2}
    \textbf{I}^{*}_n=\textbf{I}_n \cup \{\mathbf{P_k}(y_m|\textbf{I}_n)\},k=1,...,K,m=1,...,M
\end{equation}
The new training data for the second layer of ensemble will be given as follows:
\begin{equation}
    \label{eq:new_training_data}
    \textbf{D}^{*}=\{\textbf{I}^{*}_n,\textbf{Y}_n\}_{n=1}^{N}
\end{equation}


For second layer of the ensemble, we train $\{\mathcal{K}_k\}_{k=1}^{K}$ on  $\textbf{D}^{*}$ to get trained segmentation models $\{\textbf{h}^{*}_k\}_{k=1}^K$. We then need to train a combiner $\mathbf{C}$ to combine the trained models $\hat{\textbf{h}}=\mathbf{C}(\{\textbf{h}^{*}_k\}_{k=1}^K)$ for final decision making. The training of combiner will conduct on the predictions for all pixels of training images in $\textbf{D}^{*}$. Once again, the new training data $\textbf{D}^{*}$ is divided into disjoint parts $\{\textbf{D}^{*}_1, \textbf{D}^{*}_2, ..., \textbf{D}^{*}_T \}$. Then for each part $\textbf{D}^{*}_t(t=1,...,T)$, the segmentation algorithms $\{\mathcal{K}_k\}_{k=1}^{K}$ will learn on $\textbf{D}^{*} \setminus \textbf{D}^{*}_t$ to obtain segmentation models $\textbf{h}^{*}_{k,t}$. These models will then predict on $\textbf{D}^{*}_t$. The second-layer probability prediction for all images in $\textbf{D}^{*}$ is given as follows:
        
        
        
        
        
        
        

\begin{equation}
    \resizebox{0.9\columnwidth}{!}{%
    $\textbf{L}^{*} = \begin{bmatrix}
        P_1(y_1 | \textbf{I}^{*}_1(1,1)) & P_1(y_2 | \textbf{I}^{*}_1(1,1)) & \cdots & P_K(y_M | \textbf{I}^{*}_1(1,1)) \\
        
        P_1(y_1 | \textbf{I}^{*}_1(1,2)) & P_1(y_2 | \textbf{I}^{*}_1(1,2)) &  \cdots & P_K(y_M | \textbf{I}^{*}_1(1,2)) \\
        
        \cdots & \cdots & \cdots  & \cdots \\
        
        P_1(y_1 | \textbf{I}^{*}_1(W,H)) & P_1(y_2 | \textbf{I}^{*}_1(W,H)) &  \cdots & P_K(y_M | \textbf{I}^{*}_1(W,H)) \\
        
        P_1(y_1 | \textbf{I}^{*}_2(1,1)) & P_1(y_2 | \textbf{I}^{*}_2(1,1)) &  \cdots & P_K(y_M | \textbf{I}^{*}_2(1,1)) \\
        
        \cdots & \cdots & \cdots  & \cdots \\
        
        P_1(y_1 | \textbf{I}^{*}_2(W,H)) & P_1(y_2 | \textbf{I}^{*}_2(W,H)) &  \cdots & P_K(y_M | \textbf{I}^{*}_2(W,H)) \\
        
        \cdots & \cdots & \cdots  & \cdots \\
        
        P_1(y_1 | \textbf{I}^{*}_N(W,H)) & P_1(y_2 | \textbf{I}^{*}_N(W,H)) &  \cdots & P_K(y_M | \textbf{I}^{*}_N(W,H))
    \end{bmatrix}$
    }
\end{equation}
Normally, a learning algorithm trains the combiner on $\textbf{L}^{*}$ with given labels of each pixel to combine the prediction of segmentation models for the final prediction. It is noted that each row of $\textbf{L}^{*}$ is the probability predictions by $K$ segmentation models on a pixel of each training image. Therefore $\textbf{L}^{*}$ will be a matrix of $N \times W \times H$ rows and $M \times K$ columns. With a large training set and large image sizes, the size of $\textbf{L}^{*}$ will be very large. For example, on Kvasir-SEG dataset of 800 training images with image size of $(640,544)$, the matrix $\textbf{L}^{*}$ will have $800*640*544=278528000$ rows. The large size of $\textbf{L}^{*}$ causes a challenge for conventional machine learning algorithms to train the combiner on all data at once. In this paper, we use a weight-based combining method on the segmentation algorithms $\{\textbf{h}^{*}_k\}_{k=1}^K$, in which each segmentation algorithm has its own weight in the combiner. The weights are found via an optimization method. This approach is practical to train the combiner on the whole $\textbf{L}^{*}$ at once. 

\subsection{Combining method}
Let $\mathbb{W}=\{w_{k,m}\}$ be the weight matrix, in which $w_{k,m}$ is the weight associated with the segmentation model $\textbf{h}^{*}_k$ and class $y_m(k=1,...,K,m=1,...,M)$. Since the class labels of the training observations are known in advance, the weights $\mathbb{W}$ can be obtained by exploring the relationship between the second-layer probability predictions in $\textbf{L}^{*}$ and the class labels of the training pixels. The weight matrix is found by minimizing the difference between the prediction for pixel $\textbf{I}_n(i,j)$ and its true class label. From the second-layer probability prediction matrix $\textbf{L}^{*}$, we extract the probabilities associated with class $y_m$ to create matrix of size $(N \times W \times H,K)$:
        
        
        
        
        
        
        

        
        
        
        
        
        
        

\begin{equation}
    \label{eq:metadata_level_2_class_m}
    \resizebox{0.9\columnwidth}{!}{%
    $\textbf{L}^{*}_m = \begin{bmatrix}
        P_1(y_m | \textbf{I}^{*}_1(1,1)) & P_2(y_m | \textbf{I}^{*}_1(1,1)) & \cdots & P_K(y_m | \textbf{I}^{*}_1(1,1)) \\
        
        P_1(y_m | \textbf{I}^{*}_1(1,2)) & P_2(y_m | \textbf{I}^{*}_1(1,2)) &  \cdots & P_K(y_m | \textbf{I}^{*}_1(1,2)) \\
        
        \cdots & \cdots & \cdots  & \cdots \\
        
        P_1(y_m | \textbf{I}^{*}_1(W,H)) & P_2(y_m | \textbf{I}^{*}_1(W,H)) &  \cdots & P_K(y_m | \textbf{I}^{*}_1(W,H)) \\
        
        P_1(y_m | \textbf{I}^{*}_2(1,1)) & P_2(y_m | \textbf{I}^{*}_2(1,1)) &  \cdots & P_K(y_m | \textbf{I}^{*}_2(1,1)) \\
        
        \cdots & \cdots & \cdots  & \cdots \\
        
        P_1(y_m | \textbf{I}^{*}_2(W,H)) & P_2(y_m | \textbf{I}^{*}_2(W,H)) &  \cdots & P_K(y_m | \textbf{I}^{*}_2(W,H)) \\
        
        \cdots & \cdots & \cdots  & \cdots \\
        
        P_1(y_m | \textbf{I}^{*}_N(W,H)) & P_2(y_m | \textbf{I}^{*}_N(W,H)) &  \cdots & P_K(y_m | \textbf{I}^{*}_N(W,H))
    \end{bmatrix}$
    }
\end{equation}

        
        

We also define crisp label vector having size $(N \times W \times H, 1)$ associated with class $y_m$ as follows: \begin{equation}
    \label{eq:crisp_label_class_m}
    \resizebox{0.5\columnwidth}{!}{%
    $\mathbb{Y}_m = \begin{bmatrix}
        \mathbb{I}[\textbf{Y}_1(1,1)=y_m] \\
        
        \cdots \\
        
        \mathbb{I}[\textbf{Y}_1(W,H)=y_m] \\
        
        \cdots \\
        
        \mathbb{I}[\textbf{Y}_n(1,1)=y_m] \\
        
        \cdots \\
        
        \mathbb{I}[\textbf{Y}_n(W,H)=y_m] \\
        
    \end{bmatrix}$
    }
\end{equation}
where $\mathbb{I}[.]$ is the indicator function. The weight vector $\mathbb{W}_m=\{w_{k,m}\}, k=1,...,K$ of size $(K, 1)$ for class $y_m$ is then found by solving a linear regression problem:
\begin{equation}
    \label{eq:linear_regression}
    \text{min}_{\mathbb{W}_m}||\mathbf{L}^{*}_m\mathbb{W}_m-\mathbb{Y}_m||_2
\end{equation}
$\mathbb{W}_m$ can be imposed with different constraints, such as Non-Negative Least Squares, i.e. $w_{k,m} \geq 0$ \cite{lawson_square_1995, stark_square_2008}, Bounded Variable Least Squares, i.e. $l_{k,m} \leq w_{k,m} \leq u_{k,m}$ in which $l_{k,m}$ and $u_{k,m}$ are lower and upper bounds \cite{bro_constrained_1997}, respectively, and Bounded Variable with Constant Sum, i.e. $-1 < w_{k,m} < 1,\sum_{k=1}^{K}w_{k,m}=1$ \cite{zhang_sparse_2011}. In this study we simply constrain the weights between 0 and 1, i.e. $0 \leq w_{k,m} \leq 1$.  By solving $M$ different linear regression problems, we will get the optimal weight matrix $\mathbb{W}=\{\mathbb{W}_m\}_{m=1}^{M}$. 


Given an unsegmented image $\textbf{I}_{test}$, it is segmented firstly by $\{\textbf{h}_k\}_{k=1}^K$ to get the prediction matrices $\{\mathbf{P_k}(y_m|\textbf{I}_{test})\}(k=1,...,K,m=1,...,M)$. Then the augmented data is created for $\textbf{I}_{test}$  by concatenating it with $\{\mathbf{P_k}(y_m|\textbf{I}_{test})\}$ as additional image channels. 
\begin{equation}
    \label{eq:level2_test}
    \textbf{I}^{*}_{test}=\textbf{I}_{test} \cup \{\mathbf{P_k}(y_m|\textbf{I}_{test})\},k=1,...,K,m=1,...,M
\end{equation}
The trained segmentation models of the second layer $\{\textbf{h}^{*}_k\}_{k=1}^K$ are then applied on $\textbf{I}^{*}_{test}$ to get the prediction matrices $\{\mathbf{P_k}(y_m|\textbf{I}^{*}_{test})\}(k=1,...,K,m=1,...,M)$. The class memberships of an image pixel $\textbf{I}^{*}_{test}(i,j)$ are found via linear combination of the prediction probabilities and the associated weights as:
\begin{align}
    \label{eq:combine_weights}
    CM_m(\textbf{I}_{test}(i,j)) &= \sum_{k=1}^{K}w_{k,m}P_k(y_m|\textbf{I}^{*}_{test}(i,j)) \nonumber \\ 
    & = \mathbb{P}_m(\textbf{I}^{*}_{test}(i,j)) \mathbb{W}_m
\end{align}
in which $\mathbb{P}_m(\textbf{I}^{*}_{test}(i,j))$ and $\mathbb{W}_m$ are defined as follows : 
\begin{align}
    \mathbb{P}_m(\textbf{I}^{*}_{test}(i,j)) &= [P_1(y_m | \textbf{I}^{*}_{test}(i,j)), ...,  P_K(y_m | \textbf{I}^{*}_{test}(i,j))] \\
    \mathbb{W}_m &= [w_{1,m},w_{2,m},...,w_{K,m}]^T
\end{align}
Finally, the predicted class label is obtained by getting the label corresponding to the maximum value of class memberships:
        



\begin{equation}
    \label{eq:argmax_test}
    \hat{m}=argmax_{m=1,...,M}CM_m\{\textbf{I}_{test}(i,j)\}
\end{equation}
\begin{equation}
    \label{eq:which_class_test}
    \textbf{I}_{test}(i,j) \in y_{\hat{m}}
\end{equation}

The combining and training procedure is described in Algorithm \ref{algo_two_layer_ensemble}. Algorithm \ref{algo_two_layer_ensemble} receives inputs including training set $\textbf{D}=\{\textbf{I}_n,\textbf{Y}_n\}_{n=1}^N$ and segmentation algorithms $\{\mathcal{K}_k\}_{k=1}^K$. Lines 2-7 create the probability matrices via $T$-fold cross-validation procedure. Line 8 creates the augmented input data for the second layer via equations \ref{eq:level2}. Lines 10-14 create the second-level predictions for all training pixels $\textbf{L}^{*}$ via $T$-fold cross-validation procedure. Lines 16-20 find the optimal weight matrix via equation \ref{eq:linear_regression}. Lines 21-24 train the segmentation models on the original training data and the augmented data respectively. Line 25 returns the trained models and the optimal weight matrix. 

The testing procedure inputs an image $\textbf{I}_{test}$, the trained models and the optimal weight matrix (see Algorithm \ref{algo_test_two_layer_ensemble}). Lines 1-2 creates the probability matrix, while in line 3, the augmented input to the second layer is created by using equations \ref{eq:level2_test}. Lines 4-5 create second-level probability matrix from augmented input. Line 6-7 use equations \ref{eq:combine_weights},  \ref{eq:argmax_test} and \ref{eq:which_class_test} to combine the second-level predictions of segmentation models by using the optimal weight matrix $\mathbb{W}$. Finally line 8 returns the final segmentation result.

\begin{center}
\label{algo:combine_train}
\begin{algorithm}[tb]
\small
\caption{\textbf{Two-layer ensemble for segmentation}}
\begin{algorithmic}[1]\label{algo_two_layer_ensemble}
\renewcommand{\algorithmicrequire}{\textbf{Input: }}
\renewcommand{\algorithmicensure}{\textbf{Output: }}

\REQUIRE Training set $\textbf{D}=\{\textbf{I}_n,\textbf{Y}_n\}_{n=1}^N$, segmentation algorithms $\{\mathcal{K}_k\}_{k=1}^{K}$\\

\ENSURE  Trained segmentation models $\{\textbf{h}_k\}_{k=1}^K$, $\{\textbf{h}^{*}_k\}_{k=1}^K$ and optimal weights $\mathbb{W}$ \\

\STATE (Posterior probability generation)
\STATE $\{\textbf{D}_1,\textbf{D}_2,...,\textbf{D}_T\}=T-partition(\textbf{D})$
\FOR {$t \gets 1$ to $T$}
    \FOR {$k \gets 1$ to $K$}
        \STATE $\textbf{h}_{k,t}$ = $Learn(\mathcal{K}_k,\textbf{D} \setminus \textbf{D}_t)$
        \FOR {$\textbf{I}$ in $\textbf{D}_t$}
            \STATE $\{\mathbf{P_k}(y_m|\textbf{I})\}_{m=1}^{M}=Segment(\textbf{h}_{k,t},\textbf{I})$
        \ENDFOR
    \ENDFOR
\ENDFOR

\STATE $\textbf{D}^{*}=\{\textbf{I}^{*}_n,\textbf{Y}_n\}_{n=1}^N$,where $\textbf{I}^{*}_n$ is defined as equations \ref{eq:level2}

\STATE (2nd-level probability generation)
\STATE $\textbf{L}^{*}=\emptyset,\{\textbf{D}^{*}_1,\textbf{D}^{*}_2,...,\textbf{D}^{*}_T\}=T-partition(\textbf{D}^{*})$
\FOR {$t \gets 1$ to $T$}
    \FOR {$k \gets 1$ to $K$}
        \STATE $\textbf{h}^{*}_{k,t}$ = $Learn(\mathcal{K}_k,\textbf{D}^{*} \setminus \textbf{D}^{*}_t)$
        \STATE $\textbf{L}^{*}=\textbf{L}^{*} \cup $ $Segment(\textbf{h}^{*}_{k,t},\textbf{D}^{*}_t)$
    \ENDFOR
\ENDFOR

\STATE (Weight vector generation)
\FOR {$m \gets 1$ to $M$}
    \STATE Get $\textbf{L}^{*}_m$ by equation \ref{eq:metadata_level_2_class_m}
    \STATE Get $\mathbb{Y}_m$ by equation \ref{eq:crisp_label_class_m}
    \STATE Find $\mathbb{W}_m=\{w_{k,m}\},k=1,...,K$ by solving equation \ref{eq:linear_regression}  
\ENDFOR

\STATE $\mathbb{W}=\{\mathbb{W}_m\}_{m=1}^{M}$

\STATE (Base segmentation algorithms generation)
\FOR {$k \gets 1$ to $K$}
    \STATE $\textbf{h}_k = Learn(\mathcal{K}_k, \textbf{D})$
    \STATE $\textbf{h}^{*}_k = Learn(\mathcal{K}_k, \textbf{D}^{*})$
\ENDFOR


\RETURN $\{\textbf{h}_k\}_{k=1}^K$, $\{\textbf{h}^{*}_k\}_{k=1}^K$, and $\mathbb{W}$

\end{algorithmic}
\end{algorithm}
\end{center}

\begin{center}
\label{algo:test}
\begin{algorithm}[tb]
\small
\caption{\textbf{Test process for two-layer ensemble for segmentation}}
\begin{algorithmic}[1]\label{algo_test_two_layer_ensemble}
\renewcommand{\algorithmicrequire}{\textbf{Input: }}
\renewcommand{\algorithmicensure}{\textbf{Output: }}


\REQUIRE Test image $\textbf{I}_{test}$, trained segmentation models $\{\textbf{h}_k\}_{k=1}^{K},\{\textbf{h}^{*}_k\}_{k=1}^{K}$ and the weight $\mathbb{W}$\\

\ENSURE  Prediction for $\textbf{I}_{test}$ \\


\FOR {$k \gets 1$ to $K$}
    \STATE $\{\mathbf{P_k}(y_m|\textbf{I}_{test})\}_{m=1}^{M}=Segment(\textbf{h}_{k},\textbf{I}_{test})$
\ENDFOR

    
\STATE $\textbf{I}^{*}_{test}$ created from $\textbf{I}_{test}$ and $\{\mathbf{P_k}(y_m|\textbf{I}_{test})\}_{m=1}^{M}$ using \ref{eq:level2_test}


\FOR {$k \gets 1$ to $K$}
    \STATE 
    $\mathbf{P^{*}_k}(y_m|\textbf{I}_{test})$ = $\mathbf{P^{*}_k}(y_m|\textbf{I}_{test}) \cup Segment(\textbf{h}^{*}_k, \textbf{I}^{*}_{test});m=1,...,M$
\ENDFOR

    
\STATE Use \ref{eq:combine_weights} to combine the predictions $\{\mathbf{P^{*}_k}(y_m|\textbf{I}_{test})\}; m=1,...,M; k=1,...,K$ 
\STATE Use \ref{eq:argmax_test} and \ref{eq:which_class_test} to get the final prediction
\RETURN The final prediction.

\end{algorithmic}
\end{algorithm}
\end{center}
\sloppy

\section{Experimental Studies}

In this experiment, we used UNet \cite{unet_olaf_2015}, LinkNet \cite{linknet_paper_2017} and Feature Pyramid Network (FPN) \cite{fpn_paper_2017}, which are three popular segmentation architectures. The backbones used were VGG16 \cite{vgg16_paper_2015} and ResNet34  \cite{resnet_paper_2016}, pretrained on the ImageNet dataset \cite{imagenet_cvpr_2009}. In total, there were 6 segmentation models used in the experiments. All segmentation algorithms were run for 300 epochs. The number of folds in the cross-validation procedure was set to 5. We compared the performance of the proposed ensemble to the 6 segmentation algorithms and one layer ensemble system with weights-based combiner, denoted by OLE in the tables.
\subsection{Performance metrics}
The performance of our proposed method and the related benchmarks were evaluated using two popular segmentation metrics. Suppose there are $M$ classes, and there are $N$ images each having size $(W,H)$. Let $\mathbf{P}$ and $\mathbf{G}$ be the prediction of a segmentation model on these images and the corresponding ground truth:
\begin{equation}
    \label{eq:prediction_and_ground_truth}
    \mathbf{P}=[p_1,p_2,...,p_M], \mathbf{G}=[g_1,g_2,...,g_M]
\end{equation}
where $p_m$ is a vector with size $(N\times W \times H,1)$ associated with class label $y_m$ in which its element is the prediction for each pixel in the form of crisp label i.e. belonging to $\{0,1\}$. Likewise, $g_m$ is a vector with size $(N \times W \times H,1)$ associated with class label $y_m$ in which each element which is the ground truth of each pixel in the form of crisp label i.e. belonging to $\{0,1\}$. Dice coefficient for the $m^{th}$ class is then defined as follows \cite{liu_dice_2019}:
\begin{equation}
    \label{eq:dice_loss_multi_class}
    DC_m=\frac{2\mathbf{p}_m^{\mathbf{T}}\mathbf{g}_m}{||\mathbf{p}_m||^2+||\mathbf{g}_m||^2}
\end{equation}
In the context of medical image analysis, local discrepancies between contours are often of interest as well. For example, radiation treatment planning applications require quantified errors in geometric displacement to ensure target coverage, normal tissue avoidance, and similar analyses \cite{kim_bidirectional_2012}. We therefore reported one measure based on distance between geometrical contours. Let $GT_m$ and $PR_m$ be the set of coordinate vectors of the ground truth contour and prediction contour with respect to class $y_m$ respectively. The Hausdorff distance $HD$ associated with class $y_m$ is calculated as follows \cite{taha_metrics_2015} : 
\begin{equation}
    \label{eq:average_hausdorff_distance}
    HD_m=max(d(GT_m,PR_m),d(PR_m,GT_m))
\end{equation}
where $d(A,B)$ is the directed Hausdorff distance:
\begin{equation}
    \label{eq:directed_hausdorff_distance}
    d(A,B)=\frac{1}{|A|} \sum_{a \in A} \min_{b \in B}||a-b||
\end{equation}
It is noted that the low Hausdorff distance or high Dice coefficient shows the good segmentation result.
\subsection{Kvasir-SEG dataset}

The first dataset used in this paper is Kvasir-SEG \cite{noauthor_kvasir-seg_nodate}, which consists of 1000 gastrointestinal polyp images, 200 of which is used for testing. The task is to segment the polyps in the images. Comparative evaluation of the segmentation models and the proposed method in Dice coefficient and Hausdorff distance is shown in Table \ref{tab:kvasirseg_dice}. The methods having VGG16 as backbone perform poorly, with Dice measure at just 0.0. In contrast, UNet-ResNet34, LinkNet-ResNet34 and FPN-ResNet34 achieve a Dice coefficient at 0.878, 0.879 and 0.887 respectively, while OLE achieves 0.888, which is roughly the same as FPN-ResNet34. The proposed method achieves a score of 0.892, which is an increase of 0.4\% compared to the second best (OLE). For the Hausdorff distance, LinkNet-VGG16 has a very high score at around 271.7, while UNet-VGG16 achieves a score of 10.402 and FPN-VGG16 has a score of 0.0 (detect nothing).
On the other hand, among the methods using ResNet34 backbone, UNet-ResNet34 has the highest Hausdorff score at 55.591, followed by LinkNet-ResNet34 at 51.241, FPN-ResNet34 at 50.321 and OLE at 49.38 . The proposed method achieves a Hausdorff distance of 48.831, which is better than the OLE by a difference of 0.55. 

\begin{figure}
    \begin{center}
        \includegraphics[width=0.3\textwidth, height=0.3\textwidth]{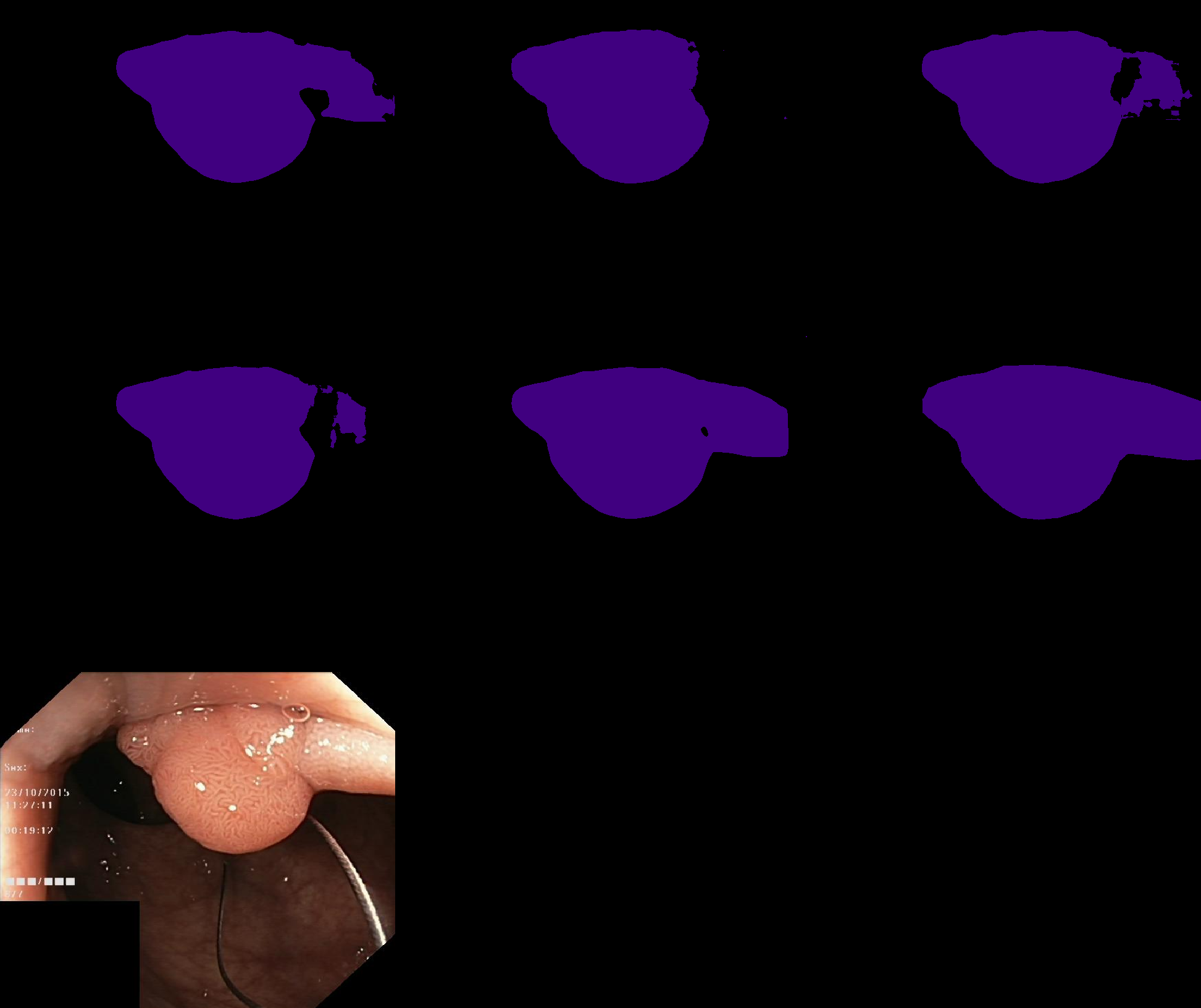}
    \end{center}
    \caption{Example result for Kvasir-SEG dataset. From left to right, top to bottom: UNet-ResNet34, LinkNet-ResNet34, FPN-ResNet34, OLE, proposed method, ground truth mask, and test image. The results made by segmentation algorithms using backbone VGG16 are not shown because they were not able to detect the polyps.}
    \label{fig:kvasir_seg_result_example_2}
\end{figure}

Figure \ref{fig:kvasir_seg_result_example_2} shows  the result of six segmentation models, OLE, the proposed ensemble, the mask of test image and the original test image. The results made by methods using backbone VGG16 are not shown because they could not predict anything. All the segmentation algorithms segmented correctly the left part of the polyp. However, for the right part, UNet-ResNet34 and FPN-ResNet34 obtained a big hole in the lower and upper part respectively, while LinkNet-ResNet34 and OLE failed to segment the right part. The proposed ensemble correctly segmented both the left and the right part of the polyp, with the exception of a relatively small hole in the middle. The reason of better performance of the proposed ensemble is that it takes into consideration information not only  from the input image but also from the  predictions in generating the segmentation models. 

The proposed ensemble has higher training time than the benchmark algorithms. Compared with OLE which took about 2 days for training on this dataset, our two-layer ensemble trained for 4 days. In our training process, we solved Equation  \ref{eq:linear_regression} to find the combining weights. Even though the optimisation problem in Equation \ref{eq:linear_regression} works on $L^{*}_m$ matrix with $278528000$ rows, it took only 5 minutes to find the weights by using \textit{sklearn} library\footnote{https://scikit-learn.org/}, which was the same as with OLE. Meanwhile, the testing time of proposed ensemble for 200 test images was 11 seconds, while OLE took 7 seconds.


\begin{table}[t]
    \centering
    \caption{Kvasir-SEG result for Dice and Hausdorff measure}
    \label{tab:kvasirseg_dice}
        \begin{tabular}{@{}lll@{}}
            \toprule
            Segmentation algorithm         & Dice  & Hausdorff      \\
            \midrule
            UNet-VGG16       & 0     & 10.402         \\
            LinkNet-VGG16    & 0.001 & 271.674        \\
            FPN-VGG16        & 0     & 0              \\
            UNet-ResNet34    & 0.878 & 55.591         \\
            LinkNet-ResNet34 & 0.879 & 51.241         \\
            FPN-ResNet34     & 0.887 & 50.321         \\
            OLE              & 0.888 & 49.38         \\
            Proposed ensemble  & \textbf{0.892} & \textbf{48.831} \\
            \bottomrule
        \end{tabular}
\end{table}

\subsection{CAMUS dataset}

\begin{figure}[ht]
    \begin{center}
        \includegraphics[width=0.3\textwidth, height=0.25\textwidth]{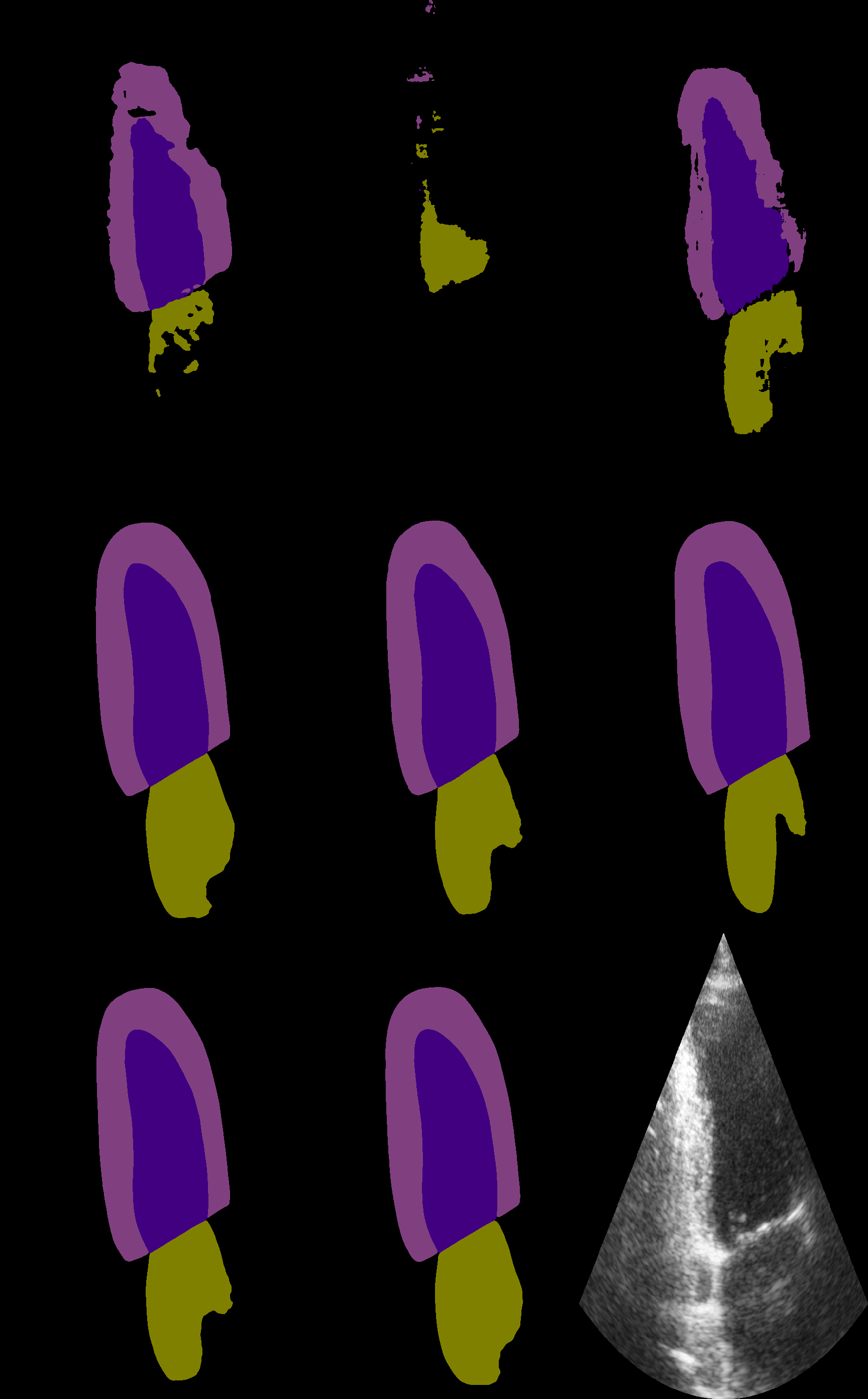}
    \end{center}
    \caption{Example result for CAMUS dataset. From left to right, top to bottom: UNet-VGG16, LinkNet-VGG16, FPN-VGG16, UNet-ResNet34, LinkNet-ResNet34, FPN-ResNet34, OLE, proposed method, and test image (mask image not available).}
    \label{fig:camus_result_good_example}
\end{figure}
The second dataset used in this paper was the Cardiac Acquisitions for Multi-structure Ultrasound Segmentation (CAMUS) dataset \cite{leclerc_deep_2019}, which is a dataset provided by a competition for accurate segmentation of 2D echocardiographic images. The dataset consists of cardiographic images and segmentation of 500 patients, acquired from clinical exams at the University Hospital of St Etienne, recorded at two cardiographic positions namely End Diastolic (ED) and End Systolic (ES). Three expert cardiologists were involved in the segmentation of the images. There are three classes: Left ventricle, Myocardium and Left atrium. The data of 50 patients are withheld for testing in which the submission link for evaluation is available here\footnote{https://www.creatis.insa-lyon.fr/Challenge/camus/scientificInterests.html}.


Table \ref{tab:result_camus_dice} and \ref{tab:result_camus_hausdorff} shows the result of the segmentation models and the proposed ensemble. We included the author's best results for each measure on this dataset \cite{leclerc_deep_2019}. It can be seen that with respect to the Dice measure, the proposed method achieved best result on all cases. For the ED case, the proposed method achieved best result on the Myocardium and Left atrium class at 0.96 and 0.907, compared to the second best result at 0.959 and 0.9 of OLE respectively. On the Left ventricle class, the proposed method achieved the same result as the second best at 0.946. For the ES case, the proposed ensemble achieved roughly the same result as OLE on Left ventricle and Myocardium class at 0.93 and 0.955 respectively. However, on Left atrium class, the proposed method achieved a score of 0.934, which is better than the second best (OLE) at 0.929. The segmentation algorithms with VGG16 backbone performed very poorly on all cases, achieving only from 0.2 (LinkNet-VGG16 on Myocardium) to 0.307 (UNet-VGG16 on Left ventricle). 

With the Hausdorff distance, the proposed ensemble beats the segmentation models in all classes for the ES case. It achieved 4.4 on the Left ventricle class while the second best among the segmentation models (LinkNet-ResNet34) achieved only 4.7 and OLE achieved 4.6. The same observation is on the Myocardium and Left atrium class. However, for the ED case, the proposed ensemble performed worse than LinkNet-VGG16, such as in the Myocardium class where the proposed method achieved a score of 5 while the LinkNet-VGG16 segmentation algorithms achieved 3.8, which is better by a score of 1.2. This can be explained from the observation in \cite{kim_bidirectional_2012} in which it is possible for the Hausdorff distance to miscalculate when the curvature has a high degree of winding and low similarity. 

Figure \ref{fig:camus_result_good_example} shows an example in which the proposed ensemble improved on the result of the segmentation models. While the predictions by the methods using VGG16 backbone (first row) contain a number of deformations compared to the test image, the predictions on the second row using ResNet34 backbone give better results. It can be seen that LinkNet-ResNet34 and FPN-ResNet34 failed to predict a large region in the bottom right of the Left atrium (second row, second and third column). On the other hand, while the prediction by UNet-ResNet34 is better than that of LinkNet-ResNet34 and FPN-ResNet34, it nevertheless contains a sharp inward region which was not correctly segmented. The proposed ensemble has improved upon the predictions by the constituent segmentation models as its prediction overall segment the bottom right part correctly.




\begin{table}[t]
\centering
\caption{Result for CAMUS dataset, Dice measure}
\label{tab:result_camus_dice}
\resizebox{0.99\columnwidth}{0.07\textheight}{
\begin{tabular}{@{}lllllll@{}}
\toprule
                                   & \multicolumn{3}{c}{End Diastolic}               & \multicolumn{3}{c}{End Systolic}  
                                   \\\cmidrule(lr){2-4}\cmidrule(lr){5-7}
                                   
                                   & Left ventricle & Myocardium    & Left atrium    & Left ventricle & Myocardium     & Left atrium    \\  \midrule
Author's best                        & 0.936          & 0.956         & 0.889          & 0.913          & 0.946          & 0.918          \\
UNet-VGG16                         & 0.307          & 0.3           & 0.244          & 0.295          & 0.305          & 0.244          \\
UNet-ResNet34                      & \textbf{0.946}          & 0.958         & 0.9            & 0.925          & 0.952          & 0.927          \\
LinkNet-VGG16                      & 0.203          & 0.2           & 0.197          & 0.106          & 0.113          & 0.119          \\
LinkNet-ResNet34                   & 0.942          & 0.958         & 0.897          & 0.928          & 0.954          & 0.922          \\
FPN-VGG16                          & 0.354          & 0.356         & 0.279          & 0.317          & 0.317          & 0.241          \\
FPN-ResNet34                       & 0.945          & 0.958         & 0.899          & 0.927          & 0.953          & 0.926          \\
OLE                       & \textbf{0.946}          & 0.959         & 0.9          & 0.929          & \textbf{0.955}          & 0.929          \\

Proposed ensemble & \textbf{0.946}          & \textbf{0.96} & \textbf{0.907} & \textbf{0.93}  & \textbf{0.955} & \textbf{0.934}          \\
\bottomrule
\end{tabular}}
\end{table}

\begin{table}[t]
\centering
\caption{Result for CAMUS dataset, Hausdorff measure}
\label{tab:result_camus_hausdorff}
\resizebox{0.99\columnwidth}{0.07\textheight}{
\begin{tabular}{@{}lllllll@{}}
\toprule
                                   & \multicolumn{3}{c}{End Diastolic}                     & \multicolumn{3}{c}{End Systolic}                               
                                   \\\cmidrule(lr){2-4}\cmidrule(lr){5-7}
                                   & Left ventricle          & Myocardium   & Left atrium  & Left ventricle          & Myocardium                        & Left atrium             \\  \midrule
Author best                        & 5.3 & 5.2          & 5.7          & 5.3 & 5.7           & 5.3 \\
UNet-VGG16                         & 15.8                    & 25.3         & 16.9         & 6.6                     & 14.3                              & 8.6                     \\
UNet-ResNet34                      & 5.1                     & 5.2          & 5            & 4.9                     & 5.3                               & 5.1                     \\
LinkNet-VGG16                      & \textbf{3.1}            & \textbf{3.8} & \textbf{4.5} & 8.1                     & 8.9                               & 9.2                     \\
LinkNet-ResNet34                   & 5                       & 5.2          & 5.5          & 4.7                     & 5.1                               & 5.5                     \\
FPN-VGG16                          & 3.8                     & 4.9          & 7.1          & 4.8                     & 6.8                               & 8                       \\
FPN-ResNet34                       & 4.8                     & 5.3          & 5.5          & 4.8                     & 5.4                               & 5.2                     \\
OLE                       & 4.7                     & 5.1          & 5.3          & 4.6                     & 5                               & 4.9                     \\
Proposed ensemble & 4.7                     & 5            & 4.8          & \textbf{4.4}            & \textbf{4.8}                      & \textbf{4.7}                     \\
\bottomrule
\end{tabular}}
\end{table}

\section{Conclusion}

In this paper, we presented a two-layer ensemble of deep learning models for segmentation of medical images. The key idea is to use the probability prediction by the constituent models in the first layer as augmented data for the second layer. The output probability prediction by the the second layer is combined by using a weight-based scheme which is not only a effective combiner but also computational efficient. The weights are found by solving a linear regression problem associated with each class label. Our results on two benchmark datasets show that the proposed ensemble method is able to combine the strengths and mitigate the drawbacks of the constituent segmentation methods, resulting in an overall improvement.
\section*{Acknowledgement}

Funding was provided by the Newton Fund Institutional Links program, project 527639907, in collaboration with Universidad Nacional Autonoma de Mexico (UNAM), granted by The British Council, UK, and Secretaría de Tecnología e Innovación (SECTEI), Mexico City, Mexico.

\bibliographystyle{IEEEtran}
\bibliography{bibfile}

\end{document}